\documentclass[lettersize,journal]{IEEEtran}
\usepackage{amsmath,amsfonts}
\usepackage{array}
\usepackage[caption=false]{subfig}
\usepackage{textcomp}
\usepackage{stfloats}
\usepackage{url}
\usepackage{verbatim}
\usepackage{graphicx}
\usepackage{cite}
\usepackage{hyperref}
\usepackage{color}
\usepackage{times}
\usepackage{epsfig}
\usepackage{graphicx}
\usepackage{amssymb}
\usepackage{float}
\usepackage{multirow}
\usepackage{diagbox}
\usepackage[ruled,linesnumbered]{algorithm2e}
\begin{document}

\title{Deep Common Feature Mining for Efficient Video Semantic Segmentation}


\author{Yaoyan~Zheng,~\IEEEmembership{Student Member,~IEEE,}
        Hongyu~Yang$^{\dag}$,~\IEEEmembership{Member,~IEEE,} 
        and Di~Huang,~\IEEEmembership{Senior Member,~IEEE,}
\IEEEcompsocitemizethanks{
\IEEEcompsocthanksitem Yaoyan Zheng is with the State Key Laboratory of Software Development Environment, School of Computer Science and Engineering, Beihang University, Beijing 100191, China (e-mail: yaoyanzheng@buaa.edu.cn).
\IEEEcompsocthanksitem Hongyu Yang is with the Institute of Artificial Intelligence, Beihang University, Beijing 100191, China, and also with Shanghai Artificial Intelligence Laboratory, Shanghai 201112, China (e-mail: hongyuyang@buaa.edu.cn).
\IEEEcompsocthanksitem Di Huang is with the State Key Laboratory of Software Development Environment, School of Computer Science and Engineering, Beihang University, Beijing 100191, China, and also with the Zhejiang Industrial Big Data and Robot Intelligent System Key Laboratory, Hangzhou Innovation Institute, Beihang University, Hangzhou 310051, China (e-mail: dhuang@buaa.edu.cn).
\IEEEcompsocthanksitem  $^{\dag}$Corresponding author.}

\thanks{Copyright © 20xx IEEE. Personal use of this material is permitted. However, permission to use this material for any other purposes must be obtained from the IEEE by sending an email to pubs-permissions@ieee.org.}
}

\markboth{IEEE TRANSACTIONS ON CIRCUITS AND SYSTEMS FOR VIDEO TECHNOLOGY}%
{Shell \MakeLowercase{\textit{Zheng et al.}}: Deep Common Feature Mining for Efficient Video Semantic Segmentation}


\maketitle
\begin{abstract}
Recent advancements in video semantic segmentation have made substantial progress by exploiting temporal correlations. Nevertheless, persistent challenges, including redundant computation and the reliability of the feature propagation process, underscore the need for further innovation. In response, we present Deep Common Feature Mining (DCFM), a novel approach strategically designed to address these challenges by leveraging the concept of feature sharing. DCFM explicitly decomposes features into two complementary components. The \textbf{common representation} extracted from a key-frame furnishes essential high-level information to neighboring non-key frames, allowing for direct re-utilization without feature propagation. Simultaneously, the \textbf{independent feature}, derived from each video frame, captures rapidly changing information, providing frame-specific clues crucial for segmentation. To achieve such decomposition, we employ a symmetric training strategy tailored for sparsely annotated data, empowering the backbone to learn a robust high-level representation enriched with common information.
Additionally, we incorporate a self-supervised loss function to reinforce intra-class feature similarity and enhance temporal consistency. Experimental evaluations on the VSPW and Cityscapes datasets demonstrate the effectiveness of our method, showing a superior balance between accuracy and efficiency. 
The implementation is available at \href{https://github.com/BUAAHugeGun/DCFM}{https://github.com/BUAAHugeGun/DCFM}.
\end{abstract}


\section{Introduction}


\IEEEPARstart{S}{emantic} segmentation is a prominent dense prediction task in the field of computer vision, involving the assignment of pixel-wise category labels to images. Substantial progress in this field has been propelled by the exploration of robust deep neural backbones \cite{resnet, segformer}, elaborate context aggregation strategies \cite{deeplabv2, prseg}, and additional boundary supervision \cite{gscnn, auxadapt}. In recent years, real-world applications like autonomous driving and robot sensing have heightened interest in video semantic segmentation (VSS). Success in VSS goes beyond high accuracy ~\cite{netwarp, the}, necessitating a trade-off between accuracy, inference efficiency \cite{lvs, dcnet} and temporal consistency \cite{etc, pc}.

\begin{figure}[t]
\centering
\includegraphics[width=\linewidth]{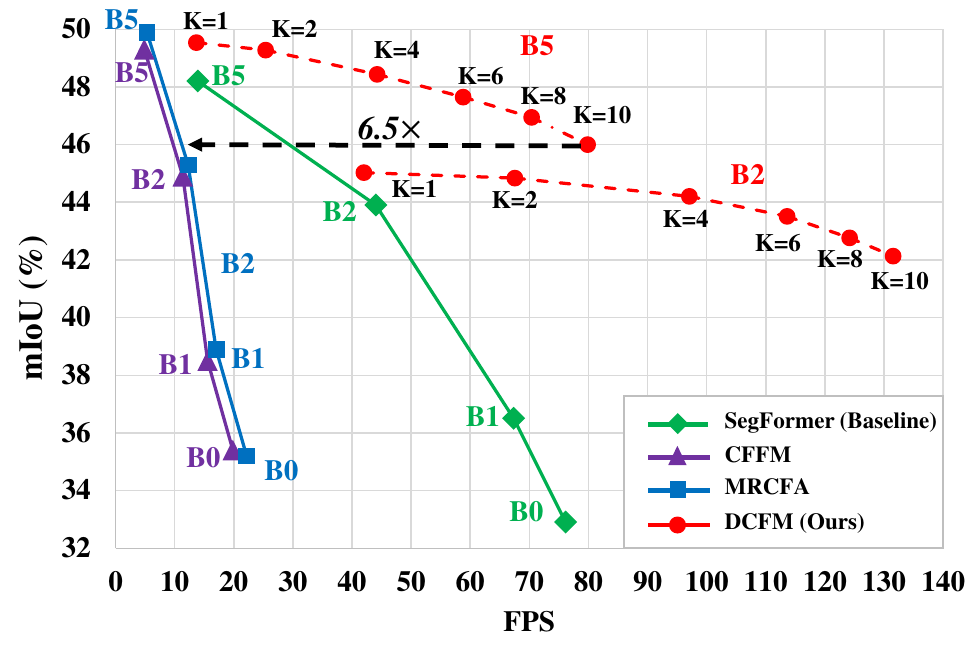}
\caption{
The trade-off between speed and accuracy is illustrated through various semantic segmentation methods using MiT backbones on the VSPW validation set \cite{vspw}. These methods include SegFormer (image baseline) \cite{segformer}, CFFM \cite{cffm}, MRCFA \cite{mrcfa}, and our proposed method, DCFM. By adjusting the keyframe interval $K$ during inference, DCFM achieves impressive speed (80 FPS @ $K$=10) while maintaining a high level of accuracy (46\% mIoU).
}
\label{fig:title}
\end{figure}

\begin{figure*}[t]
\centering
\subfloat[Per-frame inference]{
    \includegraphics[width=0.24\linewidth]{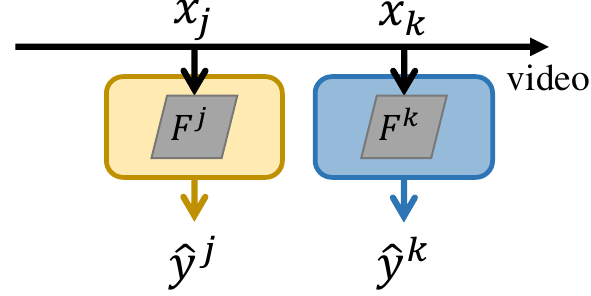}
    \label{fig:moti_a}
}
\subfloat[Feature aggregation]{
    \includegraphics[width=0.24\linewidth]{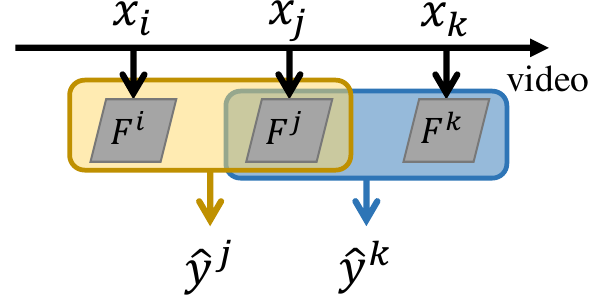}
    \label{fig:moti_b}
}
\subfloat[Feature propagation]{
    \includegraphics[width=0.24\linewidth]{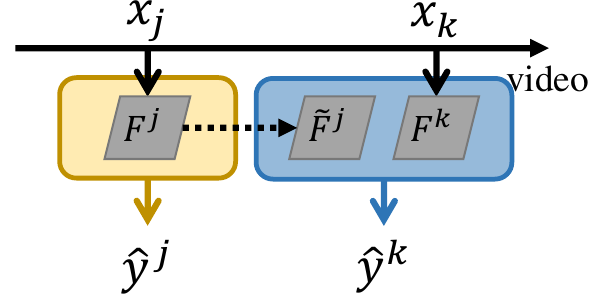}
    \label{fig:moti_c}
}
\subfloat[DCFM (Ours)]{
    \includegraphics[width=0.24\linewidth]{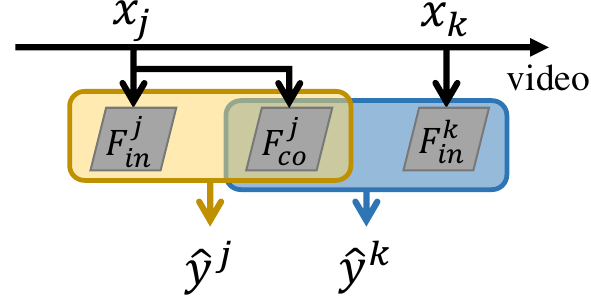}
    \label{fig:moti_d}
}

\caption{Video semantic segmentation pipelines. (a) Applying an image segmentation model to each frame independently; (b) Aggregating features from multiple frames to predict the segmentation map for a target frame, enhancing accuracy; (c) Holistically propagating the feature of a keyframe to subsequent non-key frames for reuse, improving efficiency; and (d) The proposed method decomposes the feature into two complementary parts, enabling direct reuse of the common part by non-key frames without re-calibration, improving efficiency.}
\label{fig:moti}
\end{figure*}


While it is straightforward to extend the capabilities of contemporary image models to VSS, the naive adaptation to every video frame 
introduces considerable computational overhead and neglects the inherent temporal correlations in videos, which are crucial for understanding scenes with lower uncertainty and enhancing temporal smoothness \cite{etc,pc}. To address this, methods like CFFM \cite{cffm}, MRCFA \cite{mrcfa}, and LMANet \cite{lma} explicitly aggregate information over multiple frames, utilizing Transformers \cite{attention} or spatial-temporal attention for context encoding. 
Despite their effectiveness, the heavy computational cost and complex temporal modeling hinder practical deployment \cite{netwarp, lma}. As illustrated in Figure~\ref{fig:title}, achieving high accuracy with CFFM and MRCFA in a certain experimental environment leads to a slow inference speed of approximately 6 FPS.

On the flip side, the temporal continuity in videos presents an opportunity to enhance efficiency. Acknowledging the frequent redundancy in content across adjacent frames, a promising approach is to reuse high-level features extracted from selected keyframes, striking a refined balance between inference speed and accuracy~\cite{clock, dvsn}.
This can be achieved by leveraging a pre-trained optical flow model for feature warping~\cite{flownetv2, mrdflow, asflow} or a well-designed propagation module~\cite{lvs,dcnet,ar} to transmit features or segmentation maps from keyframes to adjacent ones, significantly reducing redundant computation.
However, existing keyframe-based methods tend to excessively rely on the propagation process, leaving accurate feature prediction unsolved, especially in complex scenarios involving large motions and severe occlusions \cite{lvs, maskflow, pc}. Furthermore, propagating unstable features tends to accumulate errors \cite{vos-rpcm}.


In this study, our emphasis is on achieving efficient VSS by learning a discriminative representation to assist adjacent frames in perceiving the scene without resorting to time-consuming re-calibration operations like optical-flow-based warping. To address both aspects, we propose decomposing the entangled feature into two complementary components: \textbf{(1) the common representation}, providing essential information for segmentation and directly reusable across frames without feature propagation. This corresponds primarily to high-level semantics that remain relatively constant over time; and \textbf{(2) the independent representation}, easily extractable to offer frame-specific clues for accurate segmentation. Figure \ref{fig:moti} visualizes the distinctions between our approach and the existing ones.

To achieve this goal, we propose a novel approach called Deep Common Feature Mining (DCFM). In DCFM, we organize the deep and shallow layers of the backbone network into stages to capture common and independent representations, respectively. A lightweight fusion module 
is designed to integrate spatially misaligned multi-level features, ensuring a comprehensive representation and high efficiency.
However, a notable challenge in common feature mining arises from sparsely annotated training data, as exemplified by datasets like Cityscapes \cite{cityscapes}, where only one frame is labeled in a video. To address this limitation and facilitate the learning of a high-level feature containing sufficient common information across consecutive frames, we propose a symmetric training strategy, alternately considering labeled frames as key-frames and non-keyframes. Additionally, a self-supervised loss reinforces intra-class feature similarity, considering both static and dynamic contents.
Crucially, our model operates without re-calibration, contributing significantly to improved result robustness and achieving an extremely fast speed on non-key frames.

Our main contributions include: (1) a novel framework for efficient VSS featuring a common feature mining architecture and a symmetric training strategy; (2) a novel self-supervised consistency loss to enhance representation robustness, accommodating both static and rapidly evolving contents; and (3) substantial speed-ups over baseline models and an improved speed-accuracy trade-off compared with the current state-of-the-art.
  
\section{Related work}

\subsection{Image semantic segmentation}

Fully Convolutional Networks (FCN) \cite{fcn} paved the way for the application of Convolutional Neural Networks (CNNs) in image semantic segmentation by replacing fully connected layers with convolution layers.
Recognizing the necessity for informative spatial contexts, subsequent methods focused on refining network architectures to aggregate rich contextual information.
Among them, the encoding-decoding structures have proven to be particularly effective \cite{unet, unet++, ende, segnet, crfs}. PSPNet \cite{psp}, along with studies like DeepLabV3 \cite{deeplabv3} and OCRNet \cite{ocr}, introduced post-processing modules for enhanced spatial awareness. HRNet-V2 \cite{hrnet} utilized exchange blocks for multi-scale information fusion, while PSA \cite{psa} explored attention mechanisms within backbone blocks. 
The application of vision transformers in image tasks, as seen in Structtoken~\cite{structtoken}, SETR \cite{setr}, Graph-seg~\cite{gseg} and Segformer \cite{segformer}, further advanced this field. Additionally, MaskFormer~\cite{maskformer} and Mask2Former~\cite{mask2former} improved DETR \cite{detr} for segmentation tasks, steering away from pixel-by-pixel classification. Some methods also incorporated boundary supervision \cite{deeplab, gscnn, auxadapt, crfs} to enhance accuracy. 
In addition, a number of methods~\cite{sdanet, mic, cds, iauda, scuda} employ domain adaptation techniques to utilize additional semantic segmentation datasets, thereby aiding segmentation on the target domain.

Another notable trend aimed at designing efficient networks for practical deployment~\cite{bisenet,dfanet,dfnet,icnet,sfa}. In particular, 
BiSeNet~\cite{bisenet} aggregated feature maps from spatial and context paths to improve accuracy in lightweight networks.
STDC \cite{stdc} removed structural redundancy from BiSeNet and introduced a new backbone, considering features extracted by shallow layers as the output of the spatial path, thereby further reducing model complexity.

\subsection{Video semantic segmentation}



To exploit temporal correlations in videos, various methods \cite{netwarp, accel, pearl, grfp, td, tma, lma, cffm, mrcfa, the, mgcnet} explicitly aggregate high-level features across frames for improved accuracy. Attention mechanisms \cite{attention} and vision transformers \cite{vit}, recognized for their proficiency in long-term dependency modeling, are prominent in VSS. TDNet \cite{td} and TMANet \cite{tma}, for instance, use attention-based approaches for feature aggregation, while CFFM \cite{cffm} employs a transformer to learn a unified representation of static and motion contexts. 

Another category of methods explores the advantages of video continuity, prioritizing a favorable speed-accuracy trade-off \cite{lvs, dff, clock, dvsn, mpvss, ar, davss}. Typically, a robust baseline segmentation model is selectively applied to keyframes, and the extracted high-level features are reused by subsequent non-key frames, effectively reducing average latency. 


\label{2.TC}


The literature emphasizes temporal consistency, often achieved by aggregating multi-frame information during inference \cite{grfp, kundu2016feature, sibechi2019exploiting}. Additionally, \cite{etc} introduced a knowledge distillation-based method for consistent image semantic segmentation results, leveraging temporal loss.
To address warping errors from optical flow, \cite{pc} proposed the Perceptual Consistency (PC) measure, providing a more accurate evaluation of temporal consistency. Notably, PC served as a regularizer during training to enhance temporal consistency. The VSPW dataset \cite{vspw} contributes by providing frame-by-frame annotations, facilitating the calculation of video consistency (VC) based on ground-truth.  

\begin{figure*}[t]
\centering
\subfloat[Framework Overview]{
    \includegraphics[height=6.8cm]{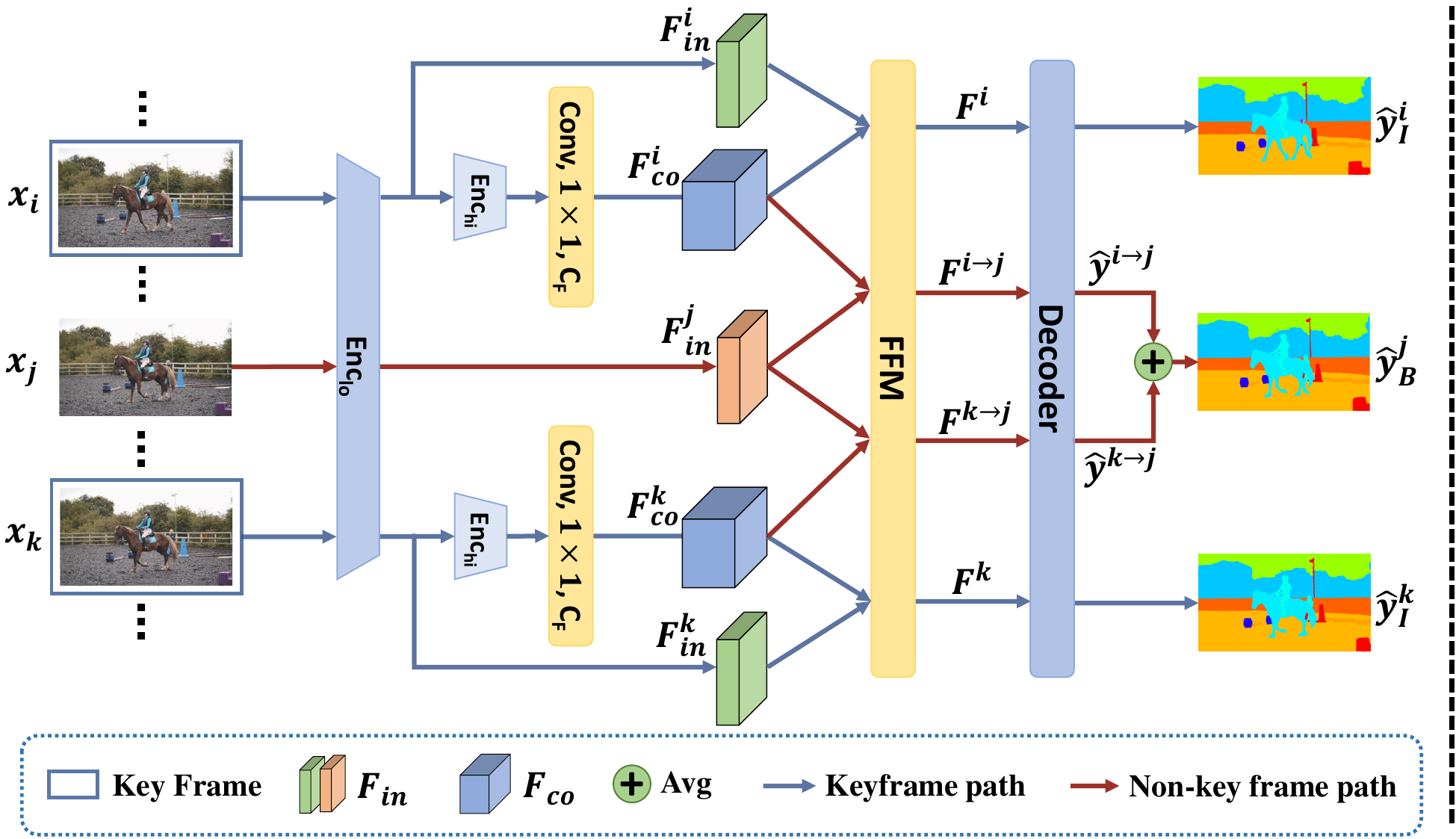}
    \label{fig:framework}
}
\hspace{-3mm}
\subfloat[Feature Fusion Module]{
    \includegraphics[height=6.8cm]{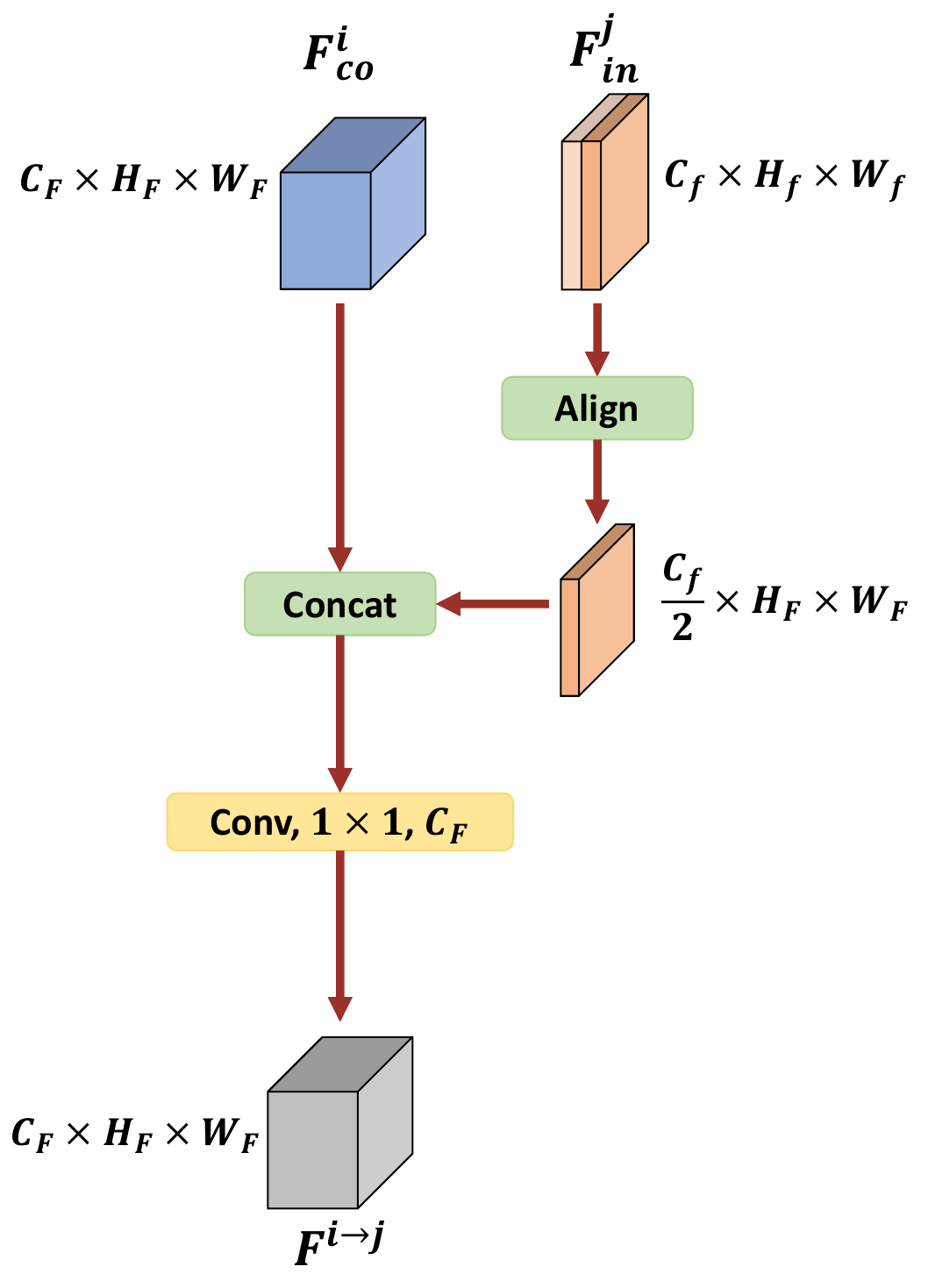}
    \label{fig:ffm}
}
\caption{Overview of Deep Common Feature Mining (DCFM) approach. (a) The complete network architecture for video segmentation during inference. The backbone, based on an image semantic segmentation network, groups layers into two stages to output common and independent features, respectively. 
The extracted common representation $F_{co}$ can be explicitly reused across frames for high efficiency. (b) Structure of the Feature Fusion Module (FFM), where the exemplar target frame is a non-key frame.}
\end{figure*}

\section{Method}

This section elaborates on the DCFM approach, detailing its network architecture, efficient inference pipeline, and the innovative training strategy incorporating a specialized loss function.

\subsection{Overview}
We employ a contemporary image segmentation model as the backbone. As observed in Clock~\cite{clock} and STDC~\cite{stdc}, features of the deep layers of the backbone tend to capture high-level semantics, evolving gradually over time. In contrast, shallow features excel at depicting details with quicker update rates. Therefore, we divide the model into two parts: $\text{Enc}(x)=\text{Enc}_{hi}(\text{Enc}_{lo}(x))$.
During training, we emphasize this property through common feature mining, ensuring deep representations from keyframes can be effortlessly reused across frames during inference, eliminating the need for re-calibration. This achieves an optimal balance between inference speed and accuracy. To accomplish this, we group the deep and shallow layers of the backbone network into two stages to capture the common and independent representations, respectively. The symmetric training strategy alternates between a labeled keyframe and a non-key frame, facilitating high-level common feature mining (detailed in Section \ref{cfm}). Additionally, a proposed self-supervised loss minimizes perceptual distance for intra-class pixels, enhancing representation robustness (detailed in Section \ref{consistency loss}).

In Figure \ref{fig:framework}, we present the complete network architecture when the model is applied to a video. For a video comprising frames $\{x_0,x_1,...,x_i,...\}$, where $x_i\in \mathbb{R}^{C\times H\times W}$, we select one keyframe for every $K$ frames. The shallow and deep layers of the backbone are organized to extract features from this keyframe $x_{i}$, expressed as:
\begin{align}
F^i_{co}=&~\text{Conv}(\text{Enc}(x_i)), \label{I-co}\\
F^i_{in}=&~\text{Enc}_{lo}(x_i), \label{I-in}
\end{align}
where $F^i_{co}\in \mathbb{R}^{C_F\times H_F\times W_F}$ and $F^i_{in}\in \mathbb{R}^{C_f\times H_f\times W_f}$ denote the deep common feature and the independent feature extracted from two certain layers of the network, respectively. Correspondingly, \textit{Enc} and $\textit{Enc}_{lo}$ represent the complete encoder and its early stage. \textit{Conv} indicates a convolution layer, mapping the encoded deep features to a fixed dimension $C_F$. The two types of spatially misaligned features are combined using a lightweight feature fusion module, denoted as \textit{FFM}, to create a more comprehensive representation, followed by a segmentation head, also known as the decoder, to produce the result $\hat{y}^i_I$:
\begin{align}
F^{i}=&~\text{FFM}(\text{Norm}(F^i_{co}), \text{Norm}(F^i_{in})), \label{I-ffm}\\
\hat{y}^{i}_I=&~\text{Dec}(F^i), \label{I-head}
\end{align}
where $F^i\in \mathbb{R}^{C_F\times H_F\times W_F}$ denotes the fused representation of the keyframe $x_i$, \textit{Norm} signifies the normalization operation, and \textit{Dec} represents the segmentation head. Similar to many previous image segmentation studies \cite{bisenet, stdc}, a combination of low-level details and high-level semantics is utilized to enhance prediction accuracy. 

Given sparse annotations in existing video segmentation datasets, we adopt a symmetric training strategy to reinforce the temporal coherence of the deep layers while ensuring that the shallow features contain rich frame-specific dynamics. At the inference stage, the independent feature $F^j_{in}$ of a non-key frame $x_j$ can be trivially extracted by a single feed-forward through the shallow layers, and the common features extracted from its two nearest keyframes, $x_{i}$ and $x_{k}$, are explicitly reused. The segmentation result $\hat{y}^{j}_B$ is efficiently obtained by:
\begin{align}
F^{i\rightarrow j}&=~\text{FFM}(\text{Norm}(F^{i}_{co}), \text{Norm}(F^j_{in})), \label{B-ffm}\\
\hat{y}^{i\rightarrow j}&=~\text{Dec}(F^{i\rightarrow j}), \label{B-head}\\
\hat{y}^{j}_B&=~\frac{1}{2}[\hat{y}^{i\rightarrow j} + \hat{y}^{k\rightarrow j}], \label{B-merge}
\end{align}
where $F^{i\rightarrow j}$ denotes the feature obtained by fusing $F^j_{in}$ with $F^{i}_{co}$, and $\hat{y}^{i\rightarrow j}$ indicates its corresponding result. Similarly, $F^{k\rightarrow j}$ and $\hat{y}^{k\rightarrow j}$ can be computed. This approach leverages both the previous and subsequent keyframes to achieve more robust segmentation results.

\begin{figure}[t]
\centering
\includegraphics[width=\linewidth]{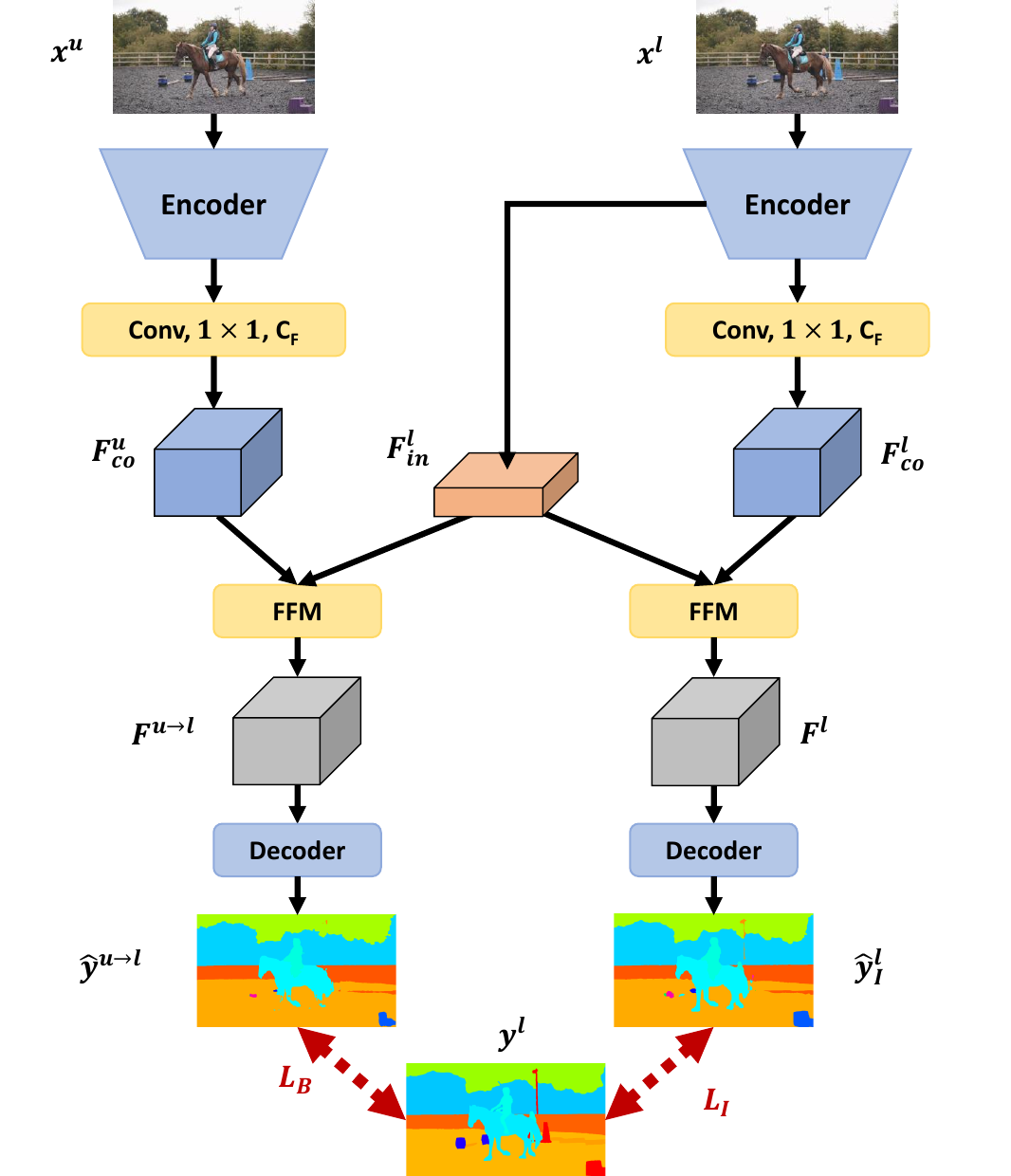}
\caption{Illustration of the common feature mining process. A labeled frame $x_l$ serves alternately as both the keyframe and non-key frame. This cyclic treatment facilitates implicit supervision of the deep features in an unlabeled neighboring frame $x_u$.}
\label{fig:cfm}
\end{figure}

\noindent\textbf{Feature fusion module.} The structure of FFM is illustrated in Figure \ref{fig:ffm}. Initially, we align the sizes of common and independent feature maps through interpolation. To alleviate computational burdens and prevent short-distance back-propagation from affecting training, we concatenate $F^i_{co}$ with half of the independent feature $F^i_{in}$ (or $F^j_{in}$). 
Then we employ only one convolution layer for feature fusion, ensuring minimal computational load. 

\subsection{Deep common feature mining}\label{cfm}

To extract common features between frames, we implement a symmetric training scheme guiding the backbone to learn an effective high-level representation with abundant common segmentation clues from adjacent frames. As illustrated in Figure \ref{fig:cfm}, each training iteration involves two neighboring frames, $x_l$ and $x_u$, where $x_l$ is labeled, and $u\in \{l-1,l+1\}$. Initially, the labeled frame $x_l$ serves as the keyframe, and 
$\hat{y}^l_I$ can be obtained by formulas \eqref{I-co}~-~\eqref{I-head}. The basic cross-entropy loss is applied, formulated as:
\begin{align}
L_I = \text{CE}(\hat{y}^l_I, y^l), \label{loss-I}
\end{align}
where $y^l$ represents the corresponding ground-truth and \textit{CE} indicates the loss function. 

The primary challenge arises from determining how to supervise the feature of the unlabeled frame. To enhance the coherence of the deep features between these two frames, we introduce implicit supervision on the unlabeled frame $F^u_{co}$ by reusing it. In particular, the labeled frame $x_l$ is repurposed as a non-key frame, while the unlabeled frame $x_u$ functions as its closest keyframe, providing the deep common representation. This allows us to obtain $\hat{y}^{u\rightarrow l}$ through the formulas \eqref{I-co}-\eqref{I-in} and \eqref{B-ffm}-\eqref{B-head}. As such, the supervision applied to the non-key frame $x_l$ can be expressed as:

\begin{align}
L_B = \text{CE}(\hat{y}^{u\rightarrow l}, y^l). \label{loss-B}
\end{align}

Minimizing the loss term (\ref{loss-B}) strengthens the consistency of common representations between neighboring frames. The alternate treatment of labeled frames as keyframes and non-key frames achieves deep common feature mining on sparsely annotated data. To facilitate learning, we further introduce intermediate supervision on the fused features, as detailed in the subsequent section.


\begin{figure}[t]
\centering
\includegraphics[width=\linewidth]{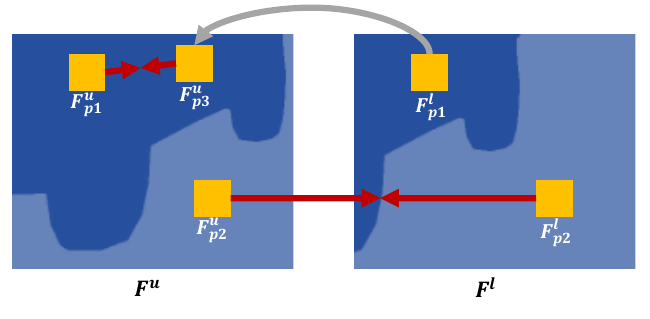}
\caption{Illustration of the proposed consistency loss. We minimize the distance between $F^u_p$ and $F^l_p$, where $p\in M_{inter}$. On a moving object, this operation is akin to minimizing the distance between $F^{u}_{p1}$ and $F^{u}_{p3}$.}
\label{fig:losspc}
\end{figure}

\subsection{Self-supervised consistency loss}

\label{consistency loss}

To alleviate the training complexity, we adopt a two-frame per-iteration training strategy. However, this choice restricts the explicit use of long-term supervisory signals to enhance the temporal consistency of segmentation results, as employed in \cite{etc,pc}. To address this limitation, we introduce a straightforward yet potent self-supervised temporal consistency loss, which can be seamlessly integrated into our framework. Specifically, this loss focuses solely on positions where predicted category labels remain constant over time, aiming to refine intra-class feature similarity. The formulation is as follows:
\begin{align}
M_{inter} =&~\mathbf{1}_{\{0\}}[\text{argmax}(\hat{y}^u_I)-\text{argmax}(\hat{y}^l_I)], \label{Mask} \\
L_{C} =&~\text{MSE}(M_{inter}\cdot F^u,M_{inter}\cdot F^l). \label{loss-PC}
\end{align}
where $M_{inter}\in \{0,1\}^{H_F\times W_F}$ denotes the intersection of two predicted semantic segmentation maps, $\hat{y}^u_I$ and $\hat{y}^l_I$.
The term $MSE$ represents Mean Square Error. Minimizing this equation effectively reduces the feature distance between the corresponding positions of the two frames, for the locations where the predicted category labels remain consistent. Importantly, we refrain from pushing features apart at positions where the predicted results change, recognizing that features at these locations are less stable.


Figure \ref{fig:losspc} visually depicts the self-supervised consistency loss $L_{C}$, which comprehensively considers both static and dynamic contents. For static objects, $L_{C}$ guides the model to learn a more stable feature representation over time, enhancing temporal consistency. For moving objects, it encourages the model to learn a more similar feature representation for pixels at different positions within the same category. Taking the car in Figure \ref{fig:losspc} as an example, the pixel at $p_{1}$ in Frame $x_{l}$ moves to $p_{3}$ in Frame $x_{u}$, and their features naturally exhibit similarity. By narrowing the feature distance between $ F^{u}_{p1}$ and $ F^{l}_{p1}$ at position $p_{1}$, the distance between $ F^{u}_{p1}$ and $ F^{u}_{p3}$ is reduced, reinforcing intra-class similarity. Overall, the self-supervised consistency loss $L_{C}$ empowers the model to acquire more robust feature representations with heightened perceptual consistency.

In summary, the joint loss is expressed as:
\begin{align}
L =L_I+\lambda_B L_B+\lambda_{C} L_{C}. \label{Loss} 
\end{align}
where $\lambda_B$ and $\lambda_{C}$ are the trade-off parameters.

\subsection{Adaptive keyframe scheduling}

Previous research \cite{lvs, dvsn, dff} has made valuable contributions to adaptive keyframe scheduling (AKS), allowing keyframe-based VSS methods to adapt to dynamic scenes with varying motion levels instead of using a fixed keyframe scheduling. 
While DCFM does not explicitly integrate a keyframe scheduling module, we use a simple AKS scheme \cite{clock} to enhance flexibility. We calculate a score between each frame and the previous keyframe during inference, defined as $s_j = \text{mean}(|x_i - x_j|)$, where $x_i$ is the previous keyframe, and $j > i$. We traverse through $x_j$ until the score $s_j$ exceeds a predefined threshold $S$ or reaches the end of the video. This adaptive keyframe selection strategy allows DCFM to adapt to varying motion levels, providing flexibility in handling dynamic scenes. 
Furthermore, to facilitate comparison with experiments utilizing a fixed keyframe interval, we introduce the concept of the minimum keyframe interval $\text{Min}_K$. We then compare various values of $\text{Min}_K$ with models employing fixed keyframe intervals in our experiments. Algorithm~\ref{alg:infer} details the inference process of DCFM, where $t_p$ and $t_s$ denote the indices of the nearest keyframes in the previous and future frames, respectively. Correspondingly, $F_p$ and $F_s$ represent the deep features of these two keyframes, and $f$ denote the shallow feature of the current frame. As shown in the algorithm, a simple AKS strategy can be easily integrated into the DCFM inference process.

\begin{algorithm}[t]
	\caption{Inference process of DCFM with a simple Adaptive Keyframe Schedulin (AKS).}
	\label{alg:infer}
	\KwIn{
Video frames $\{x_1, x_2,...,x_n\}$; 
Encoder $\text{Enc}=\text{Enc}_{hi}\circ \text{Enc}_{lo}$;
Decoder $\text{Dec}$;
Minimum keyframe interval $\text{Min}_K$;
AKS threshold $S$;
The index of the first keyframe $T$;
Feature fusion module $\text{FFM}$;
}
	\KwOut{Segmentation results: $\{\hat{y}_1,\hat{y}_2,...,\hat{y}_n\}$;}
	\BlankLine
	$t_p\gets 0$, $t_s\gets T$;
    
    $F_p\gets \text{None}$, $F_s\gets \text{Norm}(\text{Conv}(\text{Enc}(x_{t_s})))$;

    $f\gets \text{Norm}(\text{Enc}_{lo}(x_{t_s}))$;

    $\hat{y}_{t_s}\gets \text{Dec}(\text{FFM}(F_s, f))$
    
    \For{$i \gets 1$ \KwTo $n-1$}{
        \If{$i=t_s$}{
            $t_p\gets t_s$;
            
            $F_p\gets F_s$;

            $t_s\gets t_s+\text{Min}_K$;
            
            \While{\textnormal{$t_s<n$ \textbf{and} $\text{mean}(|x_{t_s}-x_{t_p}|)\le S$}}{
                $t_s\gets t_s+1$;
            }
            
            $F_s\gets \text{Norm}(\text{Conv}(\text{Enc}(x_{t_s}))$;
            
            $f \gets \text{Norm}(\text{Enc}_{lo}(x_{t_s}))$;

            $\hat{y}_{t_s}\gets \text{Dec}(\text{FFM}(F_s, f))$;
            
        }
        \Else{
            $f\gets \text{Norm}(\text{Enc}_{lo}(x_i))$;
            
            $\hat{y}_i\gets \frac{1}{2}(\text{Dec}(\text{FFM}(F_p, f))+\text{Dec}(\text{FFM}(F_s, f)))$;
        }
    }
\end{algorithm}

\begin{table}[t]
    \begin{center}
    \caption{Detailed experimental setup, where $C_{out}$ indicates the number of original feature channels produced by the backbone network. }
    \label{tab:detail}

    \resizebox{1\linewidth}{!}{
        \begin{tabular}{c|c|c|c|c}
        \hline
         Dataset & VSPW \cite{vspw} & \multicolumn{2}{c|}{Cityscapes \cite{cityscapes}} & CamVid~\cite{camvid} \\
         \cline{1-5}
         Backbone                  & MiT~\cite{segformer}& MiT~\cite{segformer} & ResNet~\cite{resnet} & ResNet~\cite{resnet} \\
        \hline\hline
        Crop size                  & $480\times 853$  & $512\times 1024$ & $768\times 1536$ & $640\times 640$\\
        Test size                  & $480\times 853$  & $512 \times 1024$& $768\times 1536$ & $720\times 960$\\
        Batch size                 & 16               & 16        & 8 & 8\\
        Optimizer                  & AdamW            & AdamW     & SGD & SGD\\
        $C_F$                      & $C_{out}$       & $C_{out}$  & $\frac{5}{16}C_{out}$ & $\frac{5}{16}C_{out}$\\
        $\lambda_B$                & 0.2              & 0.4       & 0.4 & 0.4\\
        $\lambda_C$                & 5.0              & 10.0      & 10.0 & 10.0\\
        LR (Enc)                   & 3e-5             & 6e-5      & 1e-2 & 2e-2\\
        LR (Dec)                   & 3e-4             & 6e-4      & 1e-2 & 2e-2\\
        Scheduler                  & Poly (P=1)       & Poly (P=1)& Poly (P=0.9) & Poly (P=0.9)\\
        Iterations                 & 160k             & 160k       & 80k & 40k\\
        Warm up                    & \checkmark       & \checkmark & \checkmark & $\times$\\
        Post-proc.            & $\times$         & $\times$  & PPM \cite{psp} & PPM \cite{psp}\\
        $\text{Enc}_{lo}$    & $1^{st}$ ViT block & $1^{st}$ ViT block & layer1 & layer1\\
        \hline
        
        \hline
        \end{tabular}
    }
    \end{center}
\end{table} 

\begin{table*}[t]
\caption{Comparison of accuracy (mIoU and wIoU), temporal consistency (mVC), parameter number, GFLOPs, and inference speed (FPS) on the VSPW \cite{vspw} validation dataset. Figure \ref{fig:sota-vs} provides a visual illustration, focusing on two dimensions: mIoU and FPS. $K$ indicates the keyframe interval during inference. We measure the computational costs in FLOPs (G), averaged over videos with a resolution of 480 × 853 pixels.}
    \begin{center}
        \begin{tabular}{c|c|c|c|c|c|c|c|c}
        \hline
         Methods & Backbone & mIoU(\%) $\uparrow$ & wIoU(\%) $\uparrow$ & $\text{mVC}_8$(\%) $\uparrow$ & $\text{mVC}_{16}$(\%) $\uparrow$ & Params $\downarrow$& GFLOPs $\downarrow$  & FPS $\uparrow$ \\
        \hline\hline
         DeepLabv3+ \cite{deeplabv3} & ResNet-101 & 34.7 & 58.8 & 83.2 & 78.2 & 62.7M  & 379.0 & 9.25 \\
         UperNet \cite{upernet}      & ResNet-101 & 36.5 & 58.6 & 82.6 & 76.1 & 83.2M  & 403.6 & 16.05 \\
         PSPNet \cite{psp}           & ResNet-101 & 36.5 & 58.1 & 84.2 & 79.6 & 70.5M  & 401.8 & 13.84 \\
         OCRNet \cite{ocr}           & ResNet-101 & 36.7 & 59.2 & 84.0 & 79.0 & 58.1M  & 361.7 & 14.39 \\
         MPVSS \cite{mpvss}          & Swin-T     & 39.9 & 62.0 & 85.9 & 80.4 & 114.0M & 39.7  & 32.86 \\
         MPVSS \cite{mpvss}          & Swin-S     & 40.4 & 62.0 & 86.0 & 80.7 & 108.0M & 47.3  & 30.61 \\
    Mask2Former \cite{mask2former}   & Swin-T     & 41.2 & 62.6 & 84.5 & 80.0 & 47.4M  & 114.4 & 17.13 \\
    Mask2Former \cite{mask2former}   & Swin-S     & 42.1 & 63.1 & 84.7 & 79.3 & 68.9M  & 152.2 & 14.52 \\
        \hline\hline
         SegFormer \cite{segformer}  & MiT-B0     & 32.9 & 56.8 & 82.7 & 77.3 & 3.8M   & 11.4  & 76.2 \\
         SegFormer \cite{segformer}  & MiT-B1     & 36.5 & 58.8 & 84.7 & 79.9 & 13.8M  & 21.6  & 67.3  \\
         SegFormer \cite{segformer}  & MiT-B2     & 43.9 & 63.7 & 86.0 & 81.2 & 24.8M  & 34.2  & 44.1 \\
         SegFormer \cite{segformer}  & MiT-B5     & 48.2 & 65.1 & 87.8 & 83.7 & 82.1M  & 102.3 & 13.9 \\
         \hline
         CFFM \cite{cffm}            & MiT-B2     & 44.9 & \textbf{64.9} & 89.8 & 85.8 & 26.5M  & 149.6 & 11.4 \\
         MRCFA \cite{mrcfa}          & MiT-B2     & \textbf{45.3} & 64.7 & 90.3 & 86.2 & 27.3M  & 142.0 & 12.3 \\
         DCFM (Ours, $K$=1)          & MiT-B2     & 45.0 & 64.3 & 86.6 & 82.0 & \textbf{24.8M}  & 38.9  & 42.0 \\
         DCFM (Ours, $K$=2)          & MiT-B2     & 44.8 & 64.2 & 88.1 & 83.5 & \textbf{24.8M}  & 22.9  & 67.6 \\
         DCFM (Ours, $K$=6)          & MiT-B2     & 43.5 & 63.0 & \textbf{90.8} & \textbf{86.5} & \textbf{24.8M}  & \textbf{12.3} &\textbf{113.6}\\
        \hline
         CFFM \cite{cffm}            & MiT-B5     & 49.3 & 65.8 & 90.8 & 87.1 & 85.5M  & 433.5 & 4.9 \\
         MRCFA \cite{mrcfa}          & MiT-B5     & \textbf{49.9} & \textbf{66.0} & 90.9 & 87.4 & 84.5M  & 414.4 & 5.3 \\
         DCFM (Ours, $K$=1)          & MiT-B5     & 49.5 & \textbf{66.0} & 89.0 & 85.5 & \textbf{82.1M}  & 107.0  & 13.7 \\
         DCFM (Ours, $K$=2)          & MiT-B5     & 49.3 & 65.9 & 90.2 & 86.7 & \textbf{82.1M}  & 57.0  & 25.4 \\
         DCFM (Ours, $K$=6)          & MiT-B5     & 47.6 & 64.7 &\textbf{92.3} &\textbf{88.9} & \textbf{82.1M}  & \textbf{23.6}  &\textbf{58.8} \\
        
        \hline
        \end{tabular}
    \end{center}
\label{sota-vspw}
\end{table*}

\section{Experiments}

DCFM is agnostic to specific models and seamlessly integrates with SOTA backbones. We illustrate its versatility by applying it to MiT (Transformer) and ResNet (CNNs). We conduct comprehensive evaluations on two benchmarks, namely VSPW \cite{vspw} and Cityscapes \cite{cityscapes}. Our model is compared with state-of-the-art methods, and we include ablation studies to validate the specifically designed network architecture and training loss.

\subsection{Datasets and evaluation metrics}

\textbf{VSPW.} The VSPW dataset \cite{vspw} comprises a total of 2,806 videos, distributed across training (2,806 videos), validation (343 videos), and test sets (387 videos). The dataset encompasses 124 semantic categories and has a frame rate of 15 FPS. The frame resolution is $480\times853$ pixels, providing detailed frame-by-frame annotations.

\textbf{Cityscapes.} Cityscapes \cite{cityscapes} consists of 5,000 videos, divided into training (2,975 videos), validation (500 videos), and test sets (1,525 videos). The dataset includes 19 semantic categories, and each video contains 30 frames at a frame rate of 17 FPS. The frame resolution is $1024\times2048$ pixels, with fine annotations available only on the 20$^{th}$ frame.

\textbf{CamVid.} CamVid~\cite{camvid} comprises 4 videos with a frame rate of 30 FPS. Annotations are provided for every 30th frame, covering 11 semantic categories. The dataset includes a total of 701 annotated frames, divided into 367 for training, 101 for validation, and 233 for testing. 


\textbf{Evaluation metrics.} We evaluate our approach using mean Intersection over Union (mIoU)~\cite{fcn} and frequency-weighted Intersection over Union (wIoU)~\cite{fcn} metrics on annotated frames. Additionally, we measure inference speed in frames per second (FPS). For assessing temporal consistency, we use mean Video Consistency (mVC) \cite{vspw}. For a video containing $n$ frames $\{x_1,x_2,...,x_n\}$, $\text{VC}_l$ represents the consistency of the segmentation results calculated for every $l$ adjacent frames and then averaged:

\begin{align}
\text{VC}_l = \frac{1}{n-l+1}\sum_{i=1}^{n-l+1}\frac{(\cap_i^{i+l-1}y_i)\cap(\cap_i^{i+l-1}\hat{y}_i)}{\cap_i^{i+l-1}y_i},
\end{align}
where $y_i$ and $\hat{y}_i$ represent the label and the segmentation result of the $i^{th}$ frame, respectively. Then, mVC$_l$ is obtained by averaging the VC$_l$ of each video.

\subsection{Implementation details}

We provide details of our experimental setup in Table~\ref{tab:detail}. When $C_F=C_{out}$, we eliminate the convolutional layer depicted in Figure \ref{fig:framework}, as $C_{out}$ is sufficiently small in MixTransformer (MiT)~\cite{segformer}. During training, we employ mini-batch stochastic gradient descent (SGD) and AdamW as the optimizers for the backbone networks ResNet \cite{resnet} and MiT \cite{segformer}, following the practices of PSPNet \cite{psp} and SegFormer \cite{segformer}. We implement data augmentation techniques such as random rotation at a slight angle, random resizing, random cropping, color jitter, and random horizontal flipping. The crop size is set to $480\times 480$, $768\times 1536$ and $640\times 640$ pixels for VSPW \cite{vspw}, Cityscapes \cite{cityscapes} and CamVid~\cite{camvid}, respectively. The models are trained on 4 GPUs with SyncBN \cite{synbn}, utilizing a batch size of 16, 8 and 8 for the VSPW, Cityscapes and CamVid datasets, respectively. For models using ResNet~\cite{resnet} as the backbone network, we adopt a larger crop size due to ResNet being more sensitive to input size compared to ViT models.

At the inference stage, the entire frame is fed into the model on a single Tesla V100 GPU, avoiding the sliding window input. Frame sizes are resized to $480\times 853$ for VSPW and $512\times 1024$ for Cityscapes when using MiT as the backbone. The default interval $K$ between keyframes is set to 2. In the case of inference on VSPW~\cite{vspw}, which is annotated frame-by-frame, we designate the $[(i\mod K)+1]^{th}$ frame as the first keyframe of the $i^{th}$ video. When conducting tests on Cityscapes~\cite{cityscapes} and CamVid~\cite{camvid}, we sequentially set the first keyframe of each video to frames $1$, $2$, up to $K$ and then average the results, given that only the $20^{th}$ frame is annotated. In experiments involving adaptive keyframe scheduling (AKS), we designate the $1^{st}$ frame as the initial keyframe to ensure result stability and facilitate comparisons.


\subsection{Comparison with previous approaches}

\begin{figure}[t]
\centering
\subfloat[VSPW]{
    \includegraphics[width=\linewidth]{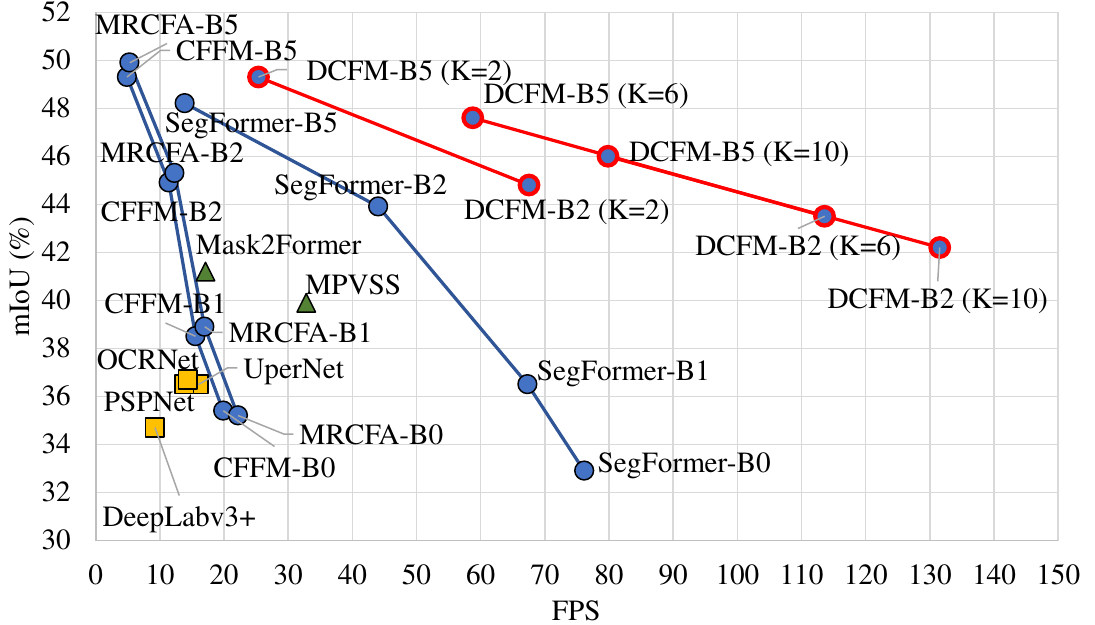}
    \label{fig:sota-vs}
}
\vspace{+0.2cm}
\subfloat[Cityscapes]{
    \includegraphics[width=\linewidth]{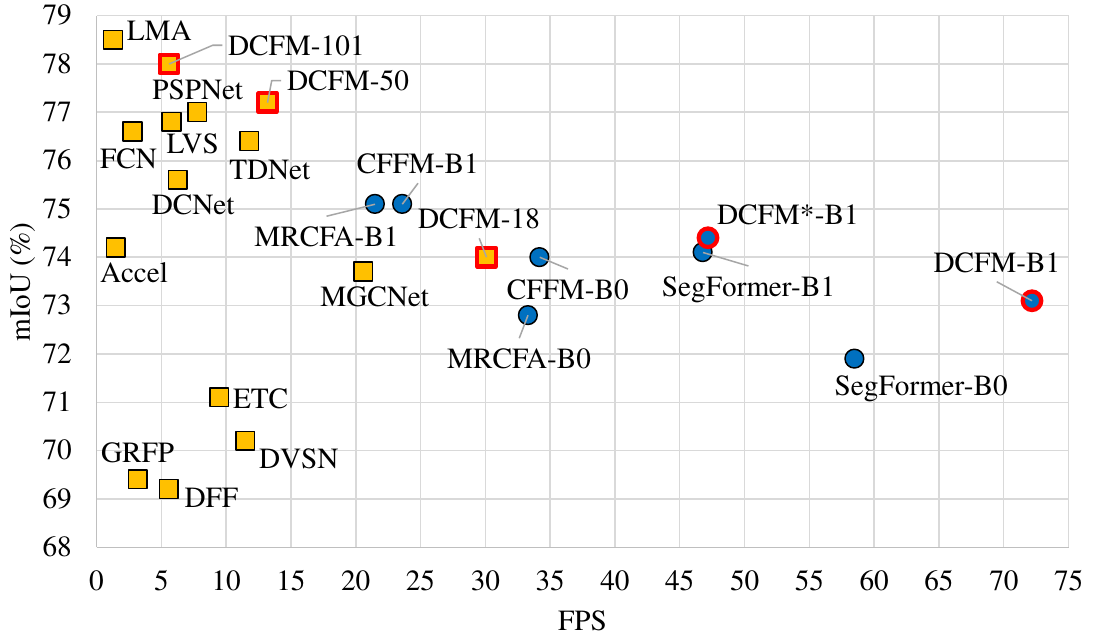}
    \label{fig:sota-cs}
}
\caption{Accuracy (mIoU) and inference speed (FPS) comparison on VSPW \cite{vspw} and Cityscapes \cite{cityscapes}.  In the figure, the yellow square, green triangle, and blue circle represent methods using ResNet \cite{resnet}, Swin \cite{swin}, and MiT \cite{segformer} as the backbone network, respectively. Our method is indicated by a red border.}
\label{fig:sota}
\end{figure}

We comprehensively compare our method with the existing ones on VSPW \cite{vspw}. Results in Table \ref{sota-vspw} highlight the notable performance of our models, particularly DCFM-B2 and DCFM-B5. Compared to the image baseline (SegFormer \cite{segformer}), DCFM-B2 and DCFM-B5 improve FPS from 44.1 and 13.9 to 67.6 and 25.4 (speed up 1.5 times and 1.8 times), respectively, while maintaining approximately 1\% higher mIoU with the keyframe interval $K$ = 2. In comparison to MRCFA-B5, the current state-of-the-art method on VSPW in terms of accuracy, DCFM-B5 ($K$ = 2) exhibits a slight decrease of -0.6\% in mIoU, but it achieves around fivefold increase in FPS (from 5.3 to 25.4). By adjusting $K$ to 6 during inference, our DCFM-B5 model achieves a significantly faster speed. Despite the slight decrease in mIoU, it still outperforms MRCFA-B2 and CFFM-B2 in terms of both accuracy and inference speed, despite having a larger parameter count. 
The results visualized in Figure \ref{fig:sota-vs} further demonstrate superior temporal consistency (mVC) and a more favorable trade-off between accuracy and efficiency achieved by the proposed approach with fewer parameters. 
Furthermore, our model tends to be easier to train due to its simple architecture without feature recalibration module as well as the proposed symmetric training scheme.

\begin{table}[t]
\caption{Performance comparison in terms of accuracy (mIoU) and inference speed (FPS) on the Cityscapes \cite{cityscapes} validation dataset. Methods, excluding ours, are arranged based on mIoU. DCFM* indicates a variant of our model with a different input resolution ($640\times 1280$). Results are visualized in Figure \ref{fig:sota-cs}.}
    \begin{center}
        \begin{tabular}{c|c|c|c}
        \hline
         Methods & Backbone & mIoU(\%) $\uparrow$ & FPS $\uparrow$ \\
        \hline\hline
        DFF \cite{dff}               & ResNet-101 & 69.2 & 5.6 \\
        GRFP \cite{grfp}             & ResNet-101 & 69.4 & 3.2 \\
        DVSN \cite{dvsn}             & ResNet-101 & 70.2 & 11.5 \\
        ETC \cite{etc}               & ResNet-18  & 71.1 & 9.5 \\
        MGC \cite{mgcnet}            & ResNet-18  & 73.7 & 20.6 \\
        Accel \cite{accel}           & ResNet-50  & 74.2 & 1.5 \\
        DCNet \cite{dcnet}           & ResNet-101 & 75.6 & 6.3 \\
        TDNet \cite{td}              & ResNet-18  & 76.4 & 11.8 \\
        FCN \cite{fcn}               & ResNet-101 & 76.6 & 2.8 \\
        LVS \cite{lvs}               & ResNet-101 & 76.8 & 5.8 \\
        PSPNet \cite{psp}            & ResNet-50  & 77.0 & 7.8 \\
        LMA \cite{lma}               & ResNet-101 & \textbf{78.5} & 1.3 \\
        DCFM (Ours)                  & ResNet-18  & 74.0 & \textbf{30.1} \\
        DCFM (Ours)                  & ResNet-50  & 77.2 & 13.2 \\
        DCFM (Ours)                  & ResNet-101 & 78.0 & 5.6 \\
        \hline\hline
        SegFormer \cite{segformer}   & MiT-B0     & 71.9 & 58.5 \\
        MRCFA \cite{mrcfa}           & MiT-B0     & 72.8 & 33.3 \\
        CFFM \cite{cffm}             & MiT-B0     & 74.0 & 34.2 \\
        SegFormer \cite{segformer}   & MiT-B1     & 74.1 & 46.8 \\
        MRCFA \cite{mrcfa}           & MiT-B1     & \textbf{75.1} & 21.5 \\
        CFFM \cite{cffm}             & MiT-B1     & \textbf{75.1} & 23.6 \\
        DCFM (Ours)                  & MiT-B1     & 73.1 & \textbf{72.2} \\
        DCFM* (Ours)                 & MiT-B1     & 74.4 & 47.2 \\
        \hline
        \end{tabular}
    \end{center}
\label{sota-cs}
\end{table}

\begin{table}[t]
\caption{Performance comparison in terms of accuracy (mIoU) and inference speed (FPS) on the CamVid \cite{camvid} test dataset. Methods, excluding ours, are arranged based on mIoU.}
    \begin{center}
        \begin{tabular}{c|c|c|c}
        \hline
         Methods & Backbone & mIoU(\%) $\uparrow$ & FPS $\uparrow$ \\
        \hline\hline
        DFF~\cite{dff}               & ResNet-101 & 66.0  & 16.1 \\
        GRFP~\cite{grfp}             & ResNet-101 & 66.1  & 4.3 \\
        Accel~\cite{accel}           & ResNet-18  & 66.7  & 7.6 \\
        TDNet~\cite{td}              & ResNet-18  & 72.6 & \textbf{25.0} \\
        DCNet~\cite{dcnet}           & ResNet-50  & 73.0 & 20.0 \\
        DCFM (Ours)                  & ResNet-50  & \textbf{73.1} & 23.7 \\
        \hline
        \end{tabular}
    \end{center}
\label{sota-cv}
\end{table} 

We extend our model comparison to the Cityscapes \cite{cityscapes}, detailed in Table \ref{sota-cs} and Figure \ref{fig:sota-cs}.
DCFM-50 outperforms contemporary feature aggregation methods like TDNet, which prioritizes high accuracy while utilizing multiple distributed networks for enhanced inference speed simultaneously. Notably, our method achieves a superior mIoU while maintaining a higher inference speed. In comparison to the method with the highest mIoU score, LMA \cite{lma}, our DCFM-101 exhibits a 0.5\% lower mIoU but is 4.3 times faster.
When compared with other keyframe-based techniques aimed at accelerating segmentation models, we surpass LVS \cite{lvs} and DCNet \cite{dcnet} in both accuracy (77.2\% mIoU) and speed (13.2 FPS) with a smaller backbone (ResNet-50 \cite{resnet}). 
These two methods also leverage different layers of the backbone network for feature reuse like ours, while our approach stands out by decoupling keyframe information into the common and independent parts.
Among the methods using MiT \cite{segformer} as the backbone, DCFM is both faster and more accurate than the SegFormer \cite{segformer} baseline. Compared with CFFM \cite{cffm} and MRCFA \cite{cffm}, the most accurate methods on VSS, DCFM achieves 0.7\% less mIoU but about twice as fast.

Additionally, we validate the effectiveness of DCFM on the CamVid~\cite{camvid} dataset. As shown in the Table~\ref{sota-cv}, DCFM achieves the highest accuracy while its FPS is only 1.3 less than the fastest method, TDNet~\cite{td}. Additionally, DCFM outperforms the latest keyframe-based method, DCNet~\cite{dcnet}, in both mIoU and FPS.

 
The results on these three datasets confirm the effectiveness of deep common feature mining, a methodology fundamentally distinct from feature aggregation over multiple frames (as shown in Figure \ref{fig:moti_b}) and keyframe-based feature propagation (as shown in Figure \ref{fig:moti_c}). In contrast to the former, mining common features across neighboring frames significantly reduces redundant computation. When compared to the latter, the decomposition operation in our model eliminates the need for unreliable feature re-calibration. Further insights into the advantage of feature decoupling are presented in the ablation experiments detailed in Section \ref{ab-decompose}.




\begin{table}[t]
\caption{Quantitative ablation study on the decoupled feature representations, achieved on the VSPW \cite{vspw} validation dataset. mIoU$_k$ and mIoU$_n$ indicate the mIoU metric calculated on keyframes and non-key frames, respectively.}
    \begin{center}
    \resizebox{\linewidth}{!}{
    \begin{tabular}{cc|cc|ccc}
    \hline
     \multicolumn{2}{c|}{Key} & \multicolumn{2}{c|}{Non-key} & \multirow{2}{*}{$\text{mIoU}_{k}$}  & \multirow{2}{*}{$\text{mIoU}_{n}$}  & \multirow{2}{*}{$\text{mIoU}_{K=2}$} \\
     \cline{1-4}
     $F^{key}_{in}$  & $F^{key}_{co}$          & $F^{non}_{in}$  & $F^{key}_{co}$               &&& \\
    \hline\hline
               & \checkmark        & \checkmark& \checkmark             & 44.2 & 43.7 & 44.0 \\
    \checkmark &                   & \checkmark& \checkmark             & 3.9  & 42.7 & 24.3 \\
    \checkmark & \checkmark        &           & \checkmark             & 44.2 & 43.4 & 43.8 \\
    \checkmark & \checkmark        & \checkmark&                        & 43.8 & 3.59 & 24.7\\
    \checkmark & \checkmark        & \checkmark& \checkmark             & \textbf{45.0} & \textbf{44.6} & \textbf{44.8}\\
    \hline
    \end{tabular}
    }
    \end{center}
\label{ab-feature}
\end{table}

\begin{table}[t]
\caption{Quantitative ablation study on the loss functions, achieved by DCFM-B2 on the VSPW~[28] validation dataset. 
}
    \begin{center}
    \begin{tabular}{ccc|c|c|c}
    \hline
     \multicolumn{3}{c|}{Loss function} & \multirow{2}{*}{mIoU (\%)} & \multirow{2}{*}{$\text{mVC}_8$(\%)} & \multirow{2}{*}{$\text{mVC}_{16}$(\%)} \\
     \cline{1-3}
     $L_I$     & $L_B$      & $L_{C}$    &  &  &     \\
    \hline\hline
    \checkmark &            &            & 43.5          & 87.4          & 82.5           \\
               & \checkmark &            & 43.0          & 87.8          & 83.3           \\
    \checkmark & \checkmark &            & 43.8          & 87.7          & 83.0           \\
    \checkmark & \checkmark & \checkmark & \textbf{45.0} & \textbf{88.1} & \textbf{83.5}  \\
    \hline
    \end{tabular}
    \end{center}
\label{tab:ab-joint}
\end{table} 

\begin{table}[t]
\caption{Quantitative ablation study using different functions as the loss function for DCFM, achieved by DCFM-B2 on the VSPW~\cite{vspw} validation dataset. 
}
    \begin{center}
    \begin{tabular}{cc|c|c|c}
    \hline
     \multicolumn{2}{c|}{Loss function} & \multirow{2}{*}{mIoU (\%)} & \multirow{2}{*}{$\text{mVC}_8$(\%)} & \multirow{2}{*}{$\text{mVC}_{16}$(\%)} \\
     \cline{1-2}
     $L_I$\&$L_B$      & $L_{C}$    &  &  &     \\
    \hline\hline
    CE+Dice~\cite{dice}     & L1         & 42.9          & \textbf{88.1} & 83.4           \\
    CE+Dice~\cite{dice}     & MSE        & 44.2          & 88.6          & 84.1           \\
    OhemCE~\cite{ohem}      & L1         & 44.5          & 88.0          & 83.4  \\
    OhemCE~\cite{ohem}      & MSE        & 43.7          & \textbf{88.2} & \textbf{83.7}  \\
    CE                      & L1         & 43.5          & 87.4          & 82.6           \\
    CE                      & MSE        & \textbf{45.0} & 88.1          & 83.5  \\
    \hline
    \end{tabular}
    \end{center}
\label{tab:ab-loss}
\end{table} 

\subsection{Ablation study}

In this section, we experimentally verify the efficacy of several designs incorporated into our model.

\textbf{Effectiveness of feature decomposing.} 
\label{ab-decompose}
The core idea of our method is to mine the common representation between frames and reuse it during inference directly. 
By decomposing features into common and independent representations, our approach distinguishes itself from other keyframe-based methods. In Table \ref{ab-feature}, we perform ablation study by omitting one of the feature components in keyframes or non-key frames during both training and testing, where MiT-B2 is exploited as the backbone.
As shown in the first row, excluding the shallow features $F^{key}_{in}$ of the keyframe causes our framework to revert to the structure of previous feature propagation-based methods, as depicted in Figure \ref{fig:moti_b}. The observed significant decrease in accuracy suggests a misalignment between the keyframes and the corresponding non-keyframes. However, instead of applying a re-calibration operation on the utilized feature to address this issue, our framework mitigates misalignments by mining common information across frames.
In the experiment of the second row, the model structure on non-key frames remains unchanged, but the accuracy mIoU$_n$ still decreases significantly. This suggests that the mining of common features and the proposed symmetric training strategy are essential.
Comparing the results of the ablation model in the third row with those obtained by the full model, it is evident that the learned common feature is highly informative, containing most of the necessary information for the non-key frames. However, utilizing the common feature $F^{key}_{co}$ provided by the keyframe alone on non-key frames is suboptimal. This is primarily because it neglects the quickly changing frame-specific dynamics, and such complementary information is well captured by the independent feature in our full model.
The third and fourth lines in the table show that using one of these two parts alone on non-key frames is not feasible. This suggests that the exceptional performance of our method on non-key frames is not solely due to the learnable network parameters but also relates to feature decomposition.

\begin{figure}[t]
\centering
\includegraphics[width=\linewidth]{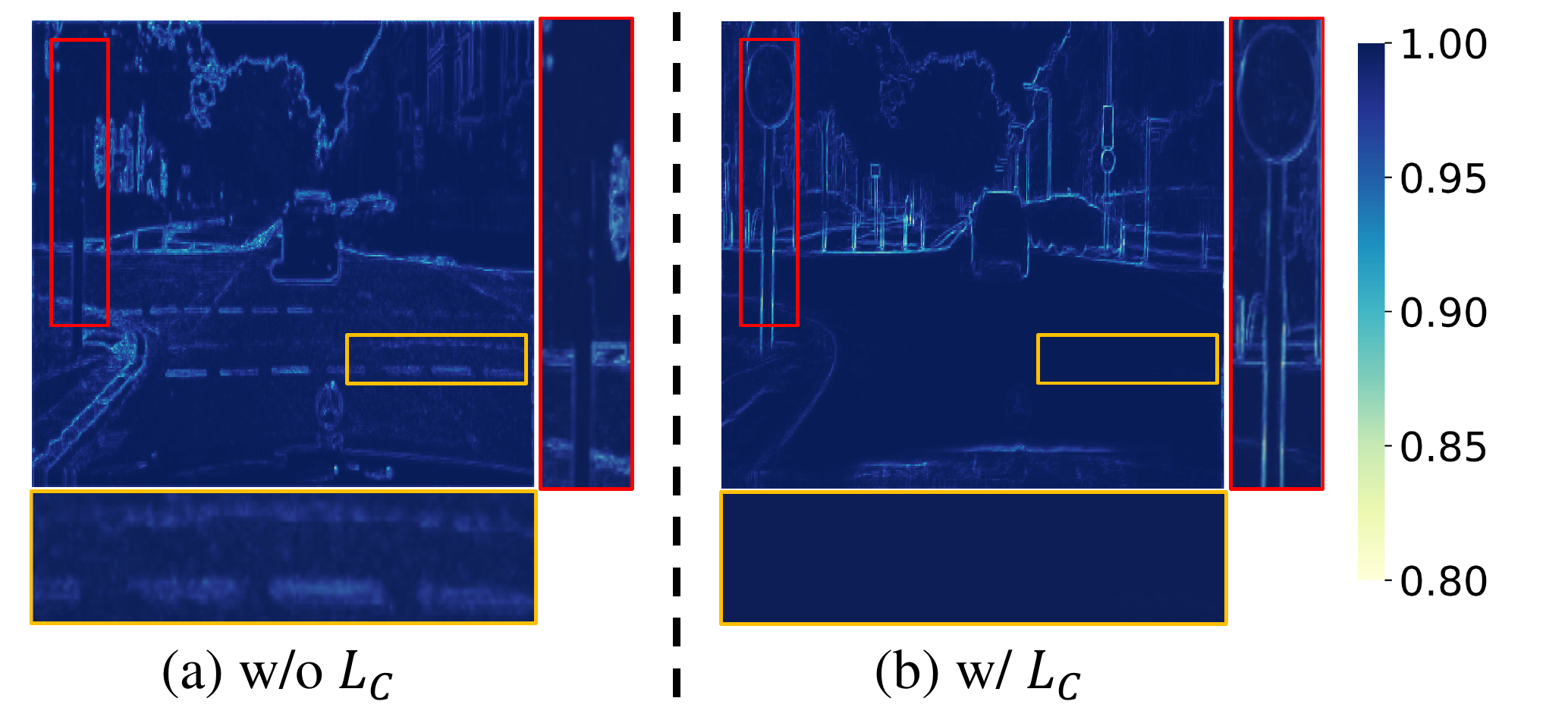}
\caption{Visualization of the cosine similarity between each pixel and its surrounding ones. 
Patches in yellow and red boxes show the effects of $L_C$ on intra- and inter-class pixels, respectively.}
\label{fig:pc}
\end{figure}

\begin{figure}[t]
\centering
\subfloat[w/o $L_{C}$]{
    \includegraphics[width=0.45\linewidth]{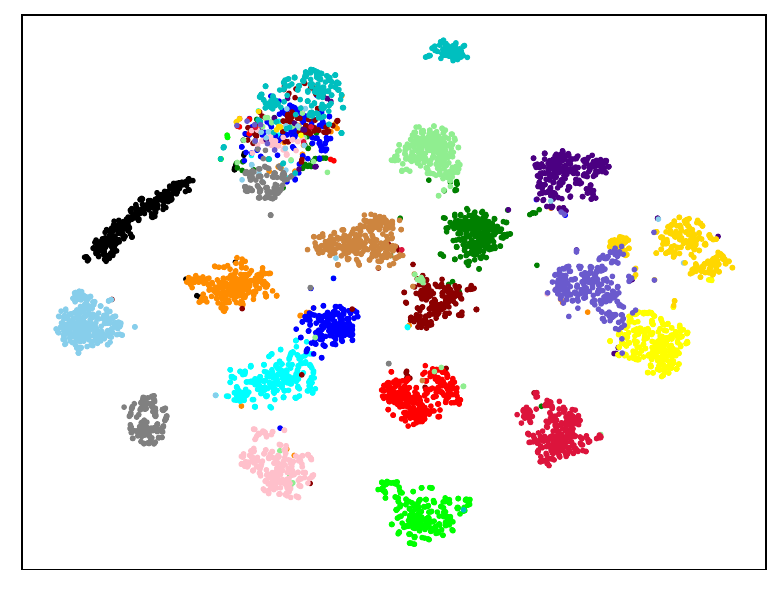}
}
\subfloat[w/ $L_{C}$]{
    \includegraphics[width=0.45\linewidth]{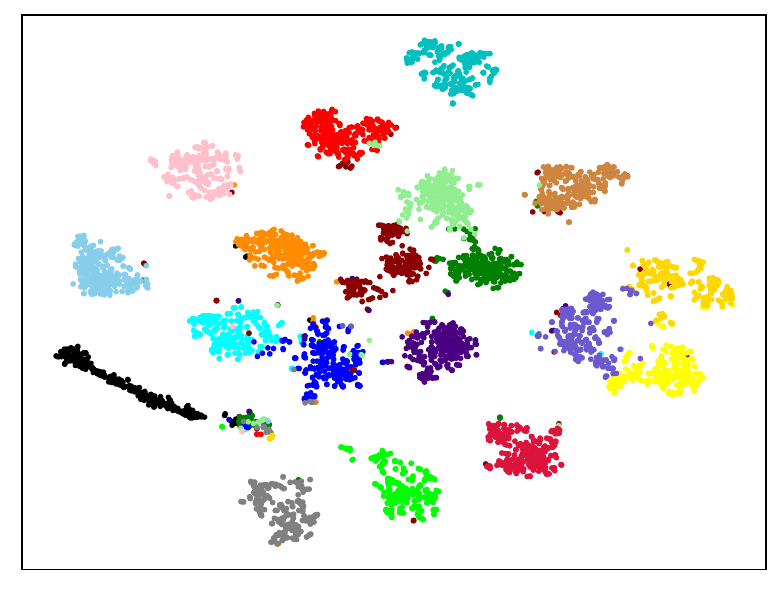}
}
\caption{t-SNE \cite{tsne} visualization of the features extracted by the models trained without (a) or with (b) $L_{C}$. Points in different colors represent feature vectors from distinct semantic categories.}
\label{fig:ab-tsne}
\end{figure}

\begin{figure}[t]
\centering
\includegraphics[width=\linewidth]{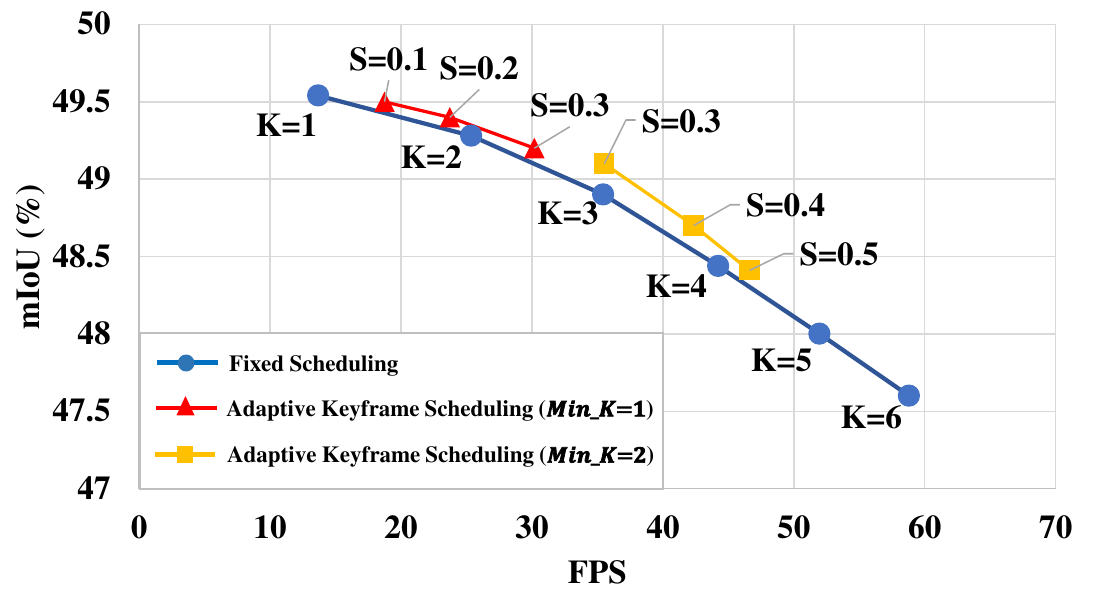}
\caption{Results of adaptive keyframe scheduling on VSPW \cite{vspw}.}
\label{fig:aks}
\end{figure}

\begin{table}[t]
\caption{Comparison of the inference speed on VSPW \cite{vspw}.}
    \begin{center}
    \resizebox{\linewidth}{!}{
        \begin{tabular}{cc|c|c|c}
        \hline
         \multicolumn{2}{c|}{\multirow{2}{*}{\diagbox[]{Modules}{Methods}}} & \multirow{2}{*}{SegFormer-B5} & \multicolumn{2}{c}{DCFM-B5 (Ours)}\\
         \cline{4-5}
                                                              & & & Key & Non-key         \\
        \hline\hline
        \multirow{2}{*}{Backbone} & $\text{Enc}_{lo}$ &2.5ms    &2.5ms     &2.5ms  \\
        \cline{2-2}
                                  & $\text{Enc}_{hi}$ &68.6ms   &68.6ms    &-  \\
        \cline{1-2}
        \multicolumn{2}{c|}{Feature Aggregation}      &-        &1.1ms     &1.6ms \\ 
        \cline{1-2}
        \multicolumn{2}{c|}{Segmentation Head}        &0.7ms    &0.7ms     &1.7ms \\
        \hline
        \multicolumn{2}{c|}{Total}                    &71.8ms   &72.9ms    &5.8ms\\
        \hline
        \end{tabular}
    }
    \end{center}
\label{ab-speed}
\end{table}

\textbf{Effectiveness of loss functions.}
Ablation studies on all training losses, as shown in Table \ref{tab:ab-joint}, validate their effectiveness. Using only $L_I$ results in the DCFM model behaving like a standard image semantic segmentation model, with unsatisfactory performance due to inadequate consideration of temporal correlations, as shown in the first row. Solely employing $L_B$ ensures consistent utilization of deep features from adjacent frames for segmentation during training. Despite a slight decrease in accuracy during testing, there is an improvement in temporal consistency, with the $\text{mVC}_{16}$ value rising from 82.5\% to 83.3\%.
Joint utilization of $L_I$ and $L_B$ improves mIoU accuracy from 43.5\% to 43.8\%.
Results in the fourth row illustrate that including $L_{C}$ significantly enhances segmentation accuracy and temporal consistency, particularly in terms of consistency over long periods of time ($\text{mVC}_{16}$). Figure \ref{fig:pc} visually evaluates the impact by calculating the cosine similarity between each pixel and its surrounding eight pixels in the feature space (calculated on the output of FFM), indicating that the learned features are less susceptible to rapidly changing details in areas with the same semantic category when $L_C$ is applied.
Furthermore, we randomly select 300 feature vectors for each category when testing DCFM on Cityscapes \cite{cityscapes} and visualize them using t-SNE \cite{tsne}. As depicted in Figure \ref{fig:ab-tsne}, the consistency loss encourages the model to learn a more robust feature representation with lower intra-class dispersion and reduced errors.

we have also conducted  ablation experiments to explore the impact of alternative loss functions such as Dice loss~\cite{dice}, OhemCE~\cite{ohem}, and L1 loss. These experiments, detailed in Table~\ref{tab:ab-loss} , reveal that while Dice loss and OhemCE lead to a slight reduction (approximately 1\%) in segmentation accuracy (mIoU), they provide minor improvements for $\text{mVC}_8$ by 0.5\% and 0.1\%. Conversely, L1 loss consistently performs worse than MSE in our context. 
These results confirm that while our framework is robust to different loss functions, MSE and cross-entropy remain effective choices for our specific task, contributing to the overall performance of the proposed method.

\textbf{Analysis on inference speed.}
We experimentally analyze the inference time of the modules in DCFM-B5 on the VSPW \cite{vspw} dataset, as shown in Table \ref{ab-speed}. The average inference time of DCFM is $\frac{(K-1)T_n+T_k}{k}$, where $T_k$ and $T_n$ respectively indicate the inference time on keyframes and non-key frames.  For the keyframe, as DCFM only uses an additional lightweight FFM module to integrate features before the segmentation head, the inference time increases by only 1.1ms compared with the SegFormer-B5 \cite{segformer} baseline, accounting for 1.5\% of the total time spent on a keyframe. Regarding the non-keyframes, the computing capacity of the adopted shallow network has only 1.48 GFLOPS, thus DCFM achieves a very fast speed of 5.8ms (172 FPS) by reusing the keyframe common features, constituting only 8\% of the keyframe inference time. As keyframe-based semantic segmentation methods, DCNet \cite{dcnet} and LVS \cite{lvs} respectively occupy 29\% and 33\% of the inference time on non-keyframes compared to keyframes. Therefore, DCFM exhibits a considerable advantage in inference speed on non-keyframes, implying that in certain fixed scenes or high frame-rate videos, DCFM can significantly reduce the required inference time by increasing the interval between keyframes while maintaining segmentation accuracy.

Additionally, as demonstrated in Table~\ref{ab-PB}, we have evaluated the performance of the proposed method when re-utilizing only previous keyframes (``P mode") in comparison to employing both previous and future keyframes (``B mode") for segmentation. The results indicate that bidirectional dependency or reliance solely on preceding keyframes does not significantly impact the overall running speed of DCFM. This efficiency is primarily attributed to the effective reuse of deep features, which incurs minimal additional computational overhead. Moreover, the flexibility to select the testing mode according to specific application scenarios further enhances the adaptability of the method.

\begin{figure*}[t]
\centering
\subfloat[VSPW]{
    \includegraphics[width=0.9\linewidth]{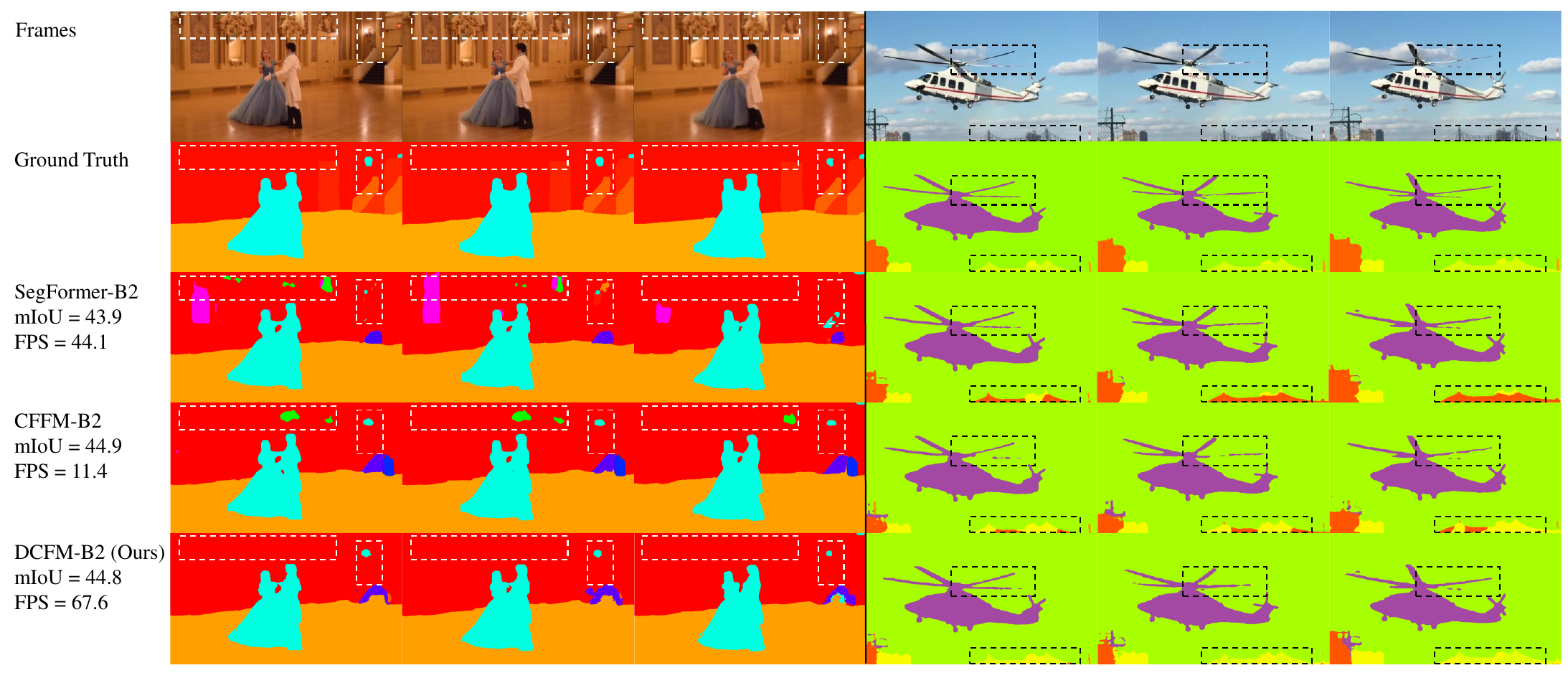}
    \label{fig:results-vspw}
    
}\\
\subfloat[Cityscapes]{
    \includegraphics[width=0.9\linewidth]{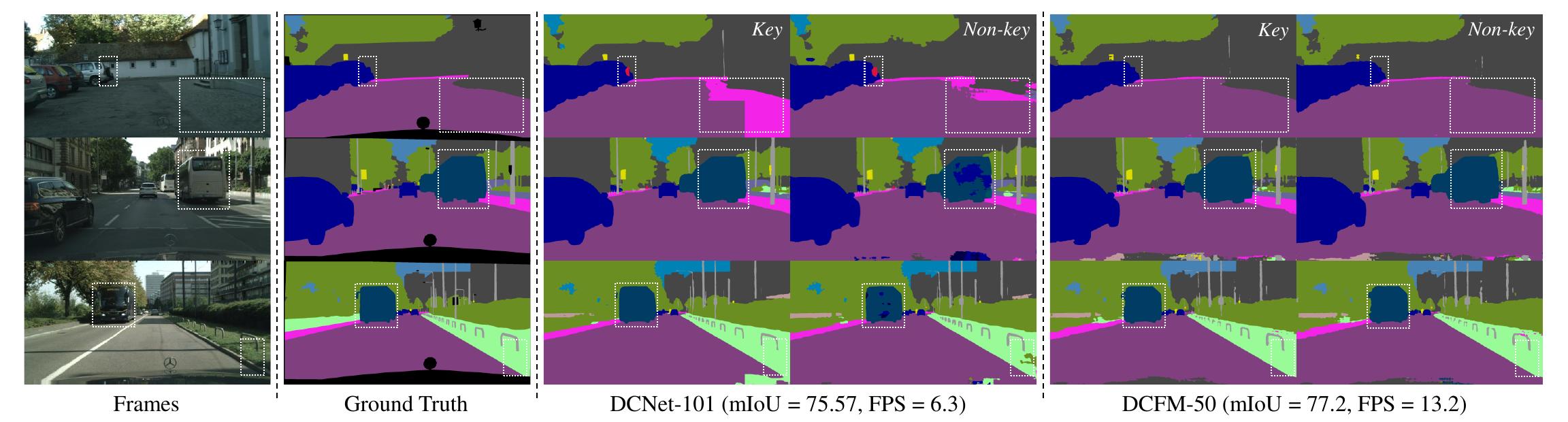}
    \label{fig:results-cs}
}
\caption{Result visualization of VSS methods on VSPW~\cite{vspw} and Cityscapes~\cite{cityscapes}. (a) Comparison of segmentation results on the VSPW~\cite{vspw} dataset with SegFormer~\cite{segformer} and CFFM~\cite{cffm}. The examples in the figure include three frames from each of two videos, demonstrating DCFM's advantages in background consistency and high-speed moving object segmentation. For DCFM, the first and third columns are segmentation results when the images are treated as non-key frames. (b) Comparison of segmentation results on the Cityscapes dataset with DCNet. The examples in the figure include three annotated frames from three different videos. The columns labeled ``Key" show the segmentation results of the two models when the frames are used as keyframes, while the columns labeled ``Kon-key" show the segmentation results when the frames are used as non-key frames, with a same distance of 1 from the nearest keyframe.}
\label{fig:results}
\end{figure*}

\textbf{Analysis on adaptive keyframe scheduling.}
As depicted in Figure \ref{fig:aks}, the accuracy and speed of DCFM-B5 exhibit consistent changes with the increase of the keyframe interval $K$. This figure also presents the results of DCFM-B5 with adaptive keyframe scheduling (AKS). For comparison, we set the maximum value of fixed interval value $K$ to be 6. For AKS, the minimum value Min$_K$ is set to 1 or 2. By adjusting the value of the threshold $S$, AKS significantly achieves a higher accuracy under a same average keyframe interval on DCFM-B5, indicating that our model performance can be further improved without re-training. 
The results also suggest that our method and the adaptive keyframe scheduling are orthogonal, allowing for the selection of specific algorithms in practical applications. In comparison with other VSS methods, AKS is excluded from the proposed DCFM. For methods such as DCNet \cite{dcnet}, LVS \cite{lvs}, and DFF \cite{dff}, we consider their best results with AKS for comparison, thereby demonstrating the effectiveness and efficiency of our approach.

\section{Result visualization}
We have performed a visual comparison of the video segmentation results on the frame-by-frame annotated VSPW~\cite{vspw} dataset with the image segmentation baseline SegFormer~\cite{segformer} and CFFM~\cite{cffm}, as shown in Figure~\ref{fig:results-vspw}. The background in the first video clearly demonstrates that DCFM outperforms SegFormer and CFFM in both spatial and temporal consistency. Observing the helicopter rotor in the second video, DCFM demonstrates superior performance in handling high-speed moving objects compared to the other methods shown. Additionally, it exhibits a significant advantage in temporal consistency for segmenting distant buildings.

Additionally, we perform a comparative evaluation between the keyframe-based representative method DCNet \cite{dcnet} and the proposed DCFM using the same experimental setup on the Cityscapes dataset \cite{cityscapes}. The segmentation results are visually illustrated in Figure~\ref{fig:results-cs}. Notably, our method demonstrates significantly improved consistency with the input frames compared to DCNet. This observation suggests that the propagation module in DCNet tends to carry unnecessary information over to non-key frames, leading to an excessive similarity between segmentation results of key and non-key frames. Specifically, the segmentation errors on non-key frames are evident, as exemplified by buses and vegetation within the white box in the figure.


\section{Conclusion}

\begin{table}[t]
\caption{Quantitative ablation study on the inference mode of non-key frames, achieved by DCFM-B2 and DCFM-b5 on the VSPW~[28] validation dataset. $P$ represents a non-key frame that only relies on the nearest previous keyframe for segmentation, while $B$ represents a non-key frame that relies on both the nearest previous and future key frames for segmentation.}
    \begin{center}
        \begin{tabular}{c|c|c|c|c}
        \hline
         Methods & Mode & mIoU(\%) & Speed (ms/f) & Latency (ms) \\
        \hline\hline
         \multirow{2}{*}{DCFM-B2} & $B$ & 44.8 & 14.8 & 90.5 \\
                                  & $P$ & 44.5 & 13.8 & 23.8  \\
        \hline
         \multirow{2}{*}{DCFM-B5} & $B$ & 49.3 & 39.4 & 136.7 \\
                                  & $P$ & 48.9 & 38.4 & 70.0  \\
        \hline
        \end{tabular}
    \end{center}
\label{ab-PB}
\end{table} 

This study addresses challenges in efficient video semantic segmentation. The proposed DCFM method aims to maintain high accuracy and temporal consistency while overcoming computational bottlenecks and feature propagation issues in existing approaches. It decomposes intricate features into two components, \textit{i.e.,} common and independent representations, where the former can be directly re-utilized by neighboring frames without additional re-calibration, contributing to an optimized trade-off between accuracy and speed. A symmetric training strategy and a self-supervised consistency loss are introduced, enhancing the ability to handle sparsely annotated data and offering potential applications in other video analysis tasks.

In contrast to existing keyframe-based VSS methods, DCFM leverages common feature mining instead of propagation modules such as optical flow, thereby enhancing computational efficiency. This efficiency is attained through the effective reuse of deep features, resulting in minimal additional computational overhead. Notably, the inference speed for non-keyframes in our method is only 8\% of that for keyframes. This design renders DCFM particularly well-suited for applications in scenarios such as security surveillance and high-resolution, high-frame-rate contexts like live video streaming, where the benefits of keyframe feature reuse can be maximized. However, similar to many keyframe-based VSS methods, the primary latency bottleneck is associated with the segmentation of keyframes, resulting in higher overall latency compared to the average segmentation time per frame. While methods like LVS~\cite{lvs} reduce latency by offloading the high-precision segmentation of keyframes to the background, real-time image semantic segmentation methods (e.g., STDC~\cite{stdc}, BiSeNet~\cite{bisenet}) are significantly more suitable for scenarios requiring low latency or low computational power~\cite{sv-rt,sv-ad}, such as autonomous driving or robotics, due to their advantages in latency and hardware optimization over VSS methods. Future work will focus on further optimizing DCFM for real-time applications and exploring additional strategies to reduce latency.

\section*{Acknowledgments}
This work is partly supported by the National Key R\&D Program of China (2022ZD0161902), the Research Program of State Key Laboratory of Software Development Environment, and the Fundamental Research Funds for the Central Universities.

\end{document}